\documentclass[letterpaper]{article} 
\usepackage{aaai24}  
\usepackage{times}  
\usepackage{helvet}  
\usepackage{courier}  
\usepackage[hyphens]{url}  
\usepackage{graphicx} 
\usepackage{natbib}  
\usepackage{caption} 
\usepackage{algorithm}
\usepackage{algorithmic}

\usepackage{amsmath} 
\usepackage{amsfonts,amssymb}
\usepackage{booktabs,makecell}
\usepackage{float}
\usepackage{multirow}
\usepackage{soul}
\usepackage[inline]{trackchanges}
\usepackage{colortbl}  
\usepackage{xcolor}
\usepackage{array}   
\definecolor{lightgray}{gray}{0.9}
\usepackage{newfloat}
\usepackage{listings}
\DeclareCaptionStyle{ruled}{labelfont=normalfont,labelsep=colon,strut=off} 
\floatstyle{ruled}
\newfloat{listing}{tb}{lst}{}
\floatname{listing}{Listing}
\title{UPDP: A Unified Progressive Depth Pruner for CNN and Vision Transformer}
\author{
    Ji Liu\equalcontrib,
    Dehua Tang\equalcontrib,
    Yuanxian Huang,
    Li Zhang,
    Xiaocheng Zeng,
    Dong Li,
    Mingjie Lu,
    Jinzhang Peng,
    Yu Wang,
    Fan Jiang,
    Lu Tian,
    Ashish Sirasao
}
\affiliations{
    Advanced Micro Devices, Inc., Beijing, China

    \{liuji, dehua.tang, yuanxian, xze, d.li, mingjiel, jinz.peng, yu.w, f.jiang, lu.tian, ashish.sirasao\}@amd.com
    
}

\usepackage{bibentry}

\begin{document}

\maketitle
\thispagestyle{plain} 
\begin{abstract}
Traditional channel-wise pruning methods by reducing network channels struggle to effectively prune efficient CNN models with depth-wise convolutional layers and certain efficient modules, such as popular inverted residual blocks. Prior depth pruning methods by reducing network depths are not suitable for pruning some efficient models due to the existence of some normalization layers. Moreover, finetuning subnet by directly removing activation layers would corrupt the original model weights, hindering the pruned model from achieving high performance. To address these issues, we propose a novel depth pruning method for efficient models. Our approach proposes a novel block pruning strategy and progressive training method for the subnet. Additionally, we extend our pruning method to vision transformer models. Experimental results demonstrate that our method consistently outperforms existing depth pruning methods across various pruning configurations. We obtained three pruned ConvNeXtV1 models with our method applying on ConvNeXtV1, which surpass most SOTA efficient models with comparable inference performance. Our method also achieves state-of-the-art pruning performance on the vision transformer model.

\end{abstract}

\section{Introduction}

Deep neural networks (DNNs) have made significant strides across various tasks, culminating in remarkable successes within industrial applications. Among these applications, the pursuit of model optimization stands out as a prevalent need, offering the potential to elevate model inference speed while minimizing accuracy trade-offs.  This pursuit encompasses a range of techniques, notably model pruning, quantization, and efficient model design. The efficient model design includes neural architecture search (NAS)~\cite{cai2020once, yu2019universally, yu2020bignas,wang2021alphanet} and handcraft design methodologies.  
\begin{figure}
    \centering   \includegraphics[width=0.48\textwidth]{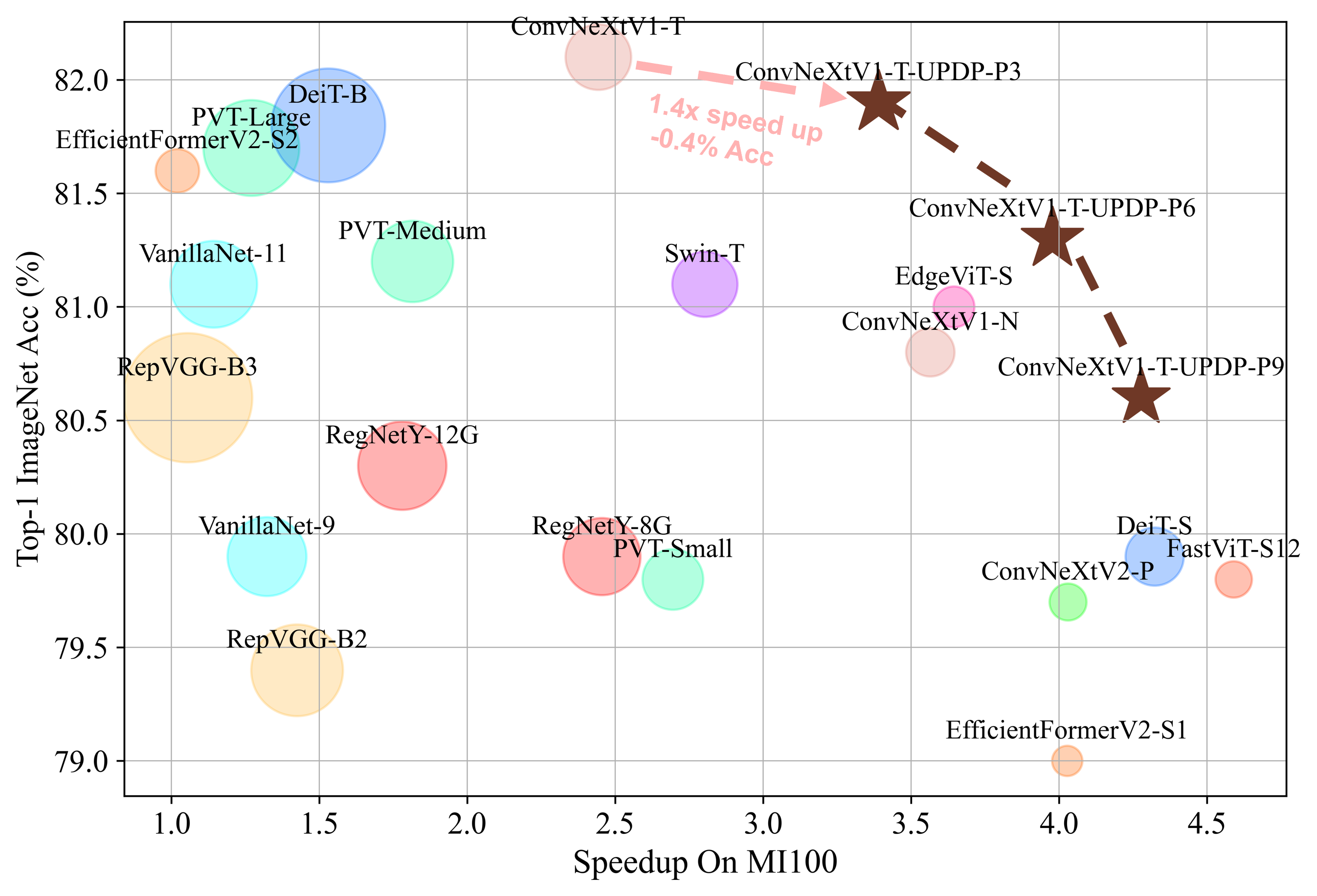}
    \caption{Performance vs. speedup on the ImageNet-1K. Our three pruned ConvNeXtV1 models surpass most SOTA efficient models on performance including RegNetY, RepVGG, VanillaNet, ConvNeXtV2, Swin-T, PVT, DeiT, EdgeViT, EfficientFormerV2, and FastViT.}
    \label{fig:sota}
\end{figure}
Model pruning has emerged as a prevalent strategy for optimizing models 
in industrial applications. Serving as a primary acceleration approach, model pruning focuses on the deliberate removal of redundant weights while maintaining accuracy. This process typically involves three sequential steps: initial baseline model training, subsequent pruning of less vital weights or layer channels, and a concluding finetuning phase for the pruned model. Notably, model pruning can be classified into two categories: non-structured pruning and structured pruning. Structured pruning is the preferred approach for model deployment in industrial applications, primarily due to hardware limitations. In contrast to non-structured methods, where less important weights in convolutional kernel layers are zeroed out in a sparse manner within each kernel channel, structured pruning encompasses techniques like channel-wise pruning and block pruning. Channel-wise pruning focuses on eliminating entire channel filters within the kernel, while block pruning operates at a larger scale, typically targeting complete blocks. Given that block pruning often leads to a reduction in model depth, it is also referred to as a depth pruner. 


The evolution of CNN model design has led to the development of more efficient models. For instance, MobileNetV2~\cite{sandler2018MobileNetV2} employs numerous depth-wise convolutional layers and stacks inverted residual blocks, achieving high performance while minimizing parameters and flops. ConvNeXtV1~\cite{liu2022convnet} leverages the large kernel trick and incorporates stacked inverted residual blocks to achieve remarkable efficiency. The conventional channel-wise pruning method faces challenges with depth-wise convolutional layers due to sparse computation and fewer parameters. 
Moreover, now model platforms favor a higher degree of parallel computing like GPUs, and channel-wise pruning methods would make efficient models thinner and sparser, which leads to low hardware utilization and thus inferior achievable hardware efficiency. To address these issues, DepthShrinker~\cite{fu2022depthshrinker} and Layer-Folding~\cite{dror2021layer} are proposed to optimize MobileNetV2 by reducing model depth through reparameterization techniques~\cite{ding2021diverse, ding2021repvgg}. 
However, these methods exhibit certain limitations. (1) The mechanism of finetuning subnet with removing activation layers directly could potentially compromise the integrity of baseline model weights, hindering the attainment of high performance. (2) These methods come with usage constraints, they are unable to prune models with some normalization layers like LayerNorm~\cite{ba2016layer} or GroupNorm~\cite{wu2018group} layer, because reparameterization technique cannot merge normalization layer which is not BatchNorm layer into adjacent convolutional layer or full-connection layer. (3) These methods cannot be applied to vision transformer models for optimization due to the existence of LayerNorm layer. 

To alleviate these problems, we propose a progressive training strategy and novel block pruning method for our depth pruning approach that can prune both CNN and vision transformer models. 
The progressive training strategy can smoothly transfer the baseline model structure to the subnet structure with high utilization of baseline model weights, which leads to higher accuracy. Our proposed block pruning method can handle the existing normalization layer issue, which can handle all activation and normalization layers in theory. Thus, our method can prune vision transformer models, which is not suitable for existing depth pruning methods.
Our experimental evaluation spans across ResNet34, MobileNetV2, and ConvNeXtV1, showcasing the superior pruning capabilities. As shown in Figure~\ref{fig:sota}, pruned ConvNeXtV1 models with our method surpass most SOTA efficient models with comparable inference performance. Notably, we extend our exploration to vision transformer models, achieving leading pruning results compared to other vision transformer pruning methods. 
   
Our main contributions can be summarized as follows.
(1) We propose a unified and efficient depth pruning method for optimizing both CNN and vision transformer models. (2) We propose a progressive training strategy for subnet optimization, coupled with a novel block pruning strategy using reparameterization technique.
(3) Conducting comprehensive experiments on both CNN and vision transformer models to showcase the superior pruning performance of our depth pruning method.
\section{Related Work}

\textbf{Network Pruning.} Pruning algorithms can be roughly divided into two types. One is the non-structured pruning algorithm represented by~\cite{han2015deep,elsen2020fast,pool2021channel}. It removes redundant elements in the weight according to certain criteria. However, non-structured pruning requires special software or hardware accelerators for the pruned models, so its versatility is not strong.  In contrast to unstructured pruning, structured pruning prunes the entire parameter structure, such as discarding entire rows or columns of weights, or entire filters in convolutional layers. 

When VGG~\cite{simonyan2014very} and ResNet~\cite{he2016deep} were on the rise, Pruning Filters~\cite{li2016pruning} adopts the L1-norm to select unimportant channels and prune them. Network 
FPGM~\cite{he2019filter} utilizes the geometric median of the convolutional filter to find redundant filters. Subsequently, various efficient DNN networks, such as MobileNet and its variants~\cite{howard2017mobilenets,howard2019searching,tan2019mnasnet,radosavovic2020designing}, incorporated depthwise convolutions~\cite{chollet2017xception} to accelerate speed and improve accuracy, enabling real-time deployment on diverse hardware platforms. 
MatePruning~\cite{liu2019metapruning} proposes the concept of PruningNet, which automatically generates weights for the pruned model, thus avoiding retraining.
However, while depthwise convolutional offers advantages in terms of reduced computation and parameters, it also presents a drawback—an increased memory footprint, posing a challenge for computationally intensive hardware like GPUs and DSPs~\cite{tan2021efficientnetv2}. Unfortunately, channel-wise pruning methods do not offer an intuitive and efficient solution to address this memory footprint challenge. 

The most relevant to our work is layer-wise pruning, which can completely remove a block or layer to reduce the depth of the network and effectively alleviate the problem of memory usage. Shallowing deep networks~\cite{chen2018shallowing} and LayerPrune~\cite{elkerdawy2020filter} propose their own strategies for evaluating the importance of convolutional layers. ESNB~\cite{zhou2021evolutionary} and ResConv~\cite{xu2020layer} identify which layers to be pruned by evolutionary search algorithms and differentiable parameters, respectively. 
Layer-Folding~\cite{dror2021layer} and DepthShinker~\cite{fu2022depthshrinker} remove non-linear activation functions within the block and merge multiple layers into a single layer using structural reparameterization techniques. Layer-Folding and DepthShinker have only been verified on the few limited models, and the hard removal of ReLU may have an impact on the accuracy of the subnet.

The Transformer family of models excels in performance across various vision tasks~\cite{carion2020end,strudel2021segmenter,brown2020language}; however, its high inference cost and significant memory footprint hinder widespread adoption~\cite{pope2023efficiently}. To tackle the memory footprint challenge, layer-wise pruning presents an effective solution. 
Dynamic skipping blocks to remove some layers has become the mainstream transformer compression method~\cite{zhang2020accelerating,dong2021attention,michel2019sixteen}. DynamicViT~\cite{rao2021dynamicvit} dynamically screens the number of tokens that need to be passed to the next layer. By encouraging dimension-wise sparsity, VTP~\cite{zhu2021vision} selects the dimension with strong redundancy for pruning.

\textbf{Structural Reparameterization}. In the absence of a nonlinear activation function within a block, the structural reparameterization technique facilitates the consolidation of multiple convolutional layers into a single convolutional layer~\cite{bhardwaj2022collapsible}. This consolidation effectively diminishes the neural network's memory requirements during inference, resulting in accelerated model processing. RepVGG~\cite{ding2021repvgg} distinguishes between training and testing structures, empowering the plain network to surpass the performance of ResNet. Furthermore, DBB~\cite{ding2021diverse} merges a multi-branch architecture into a single convolution, significantly outpacing the speed of a conventional multi-branch unit.

\textbf{Neural Architecture Search (NAS).} Weight-sharing NAS has become the mainstream of pruning methods due to its flexibility and convenience of training a supernet and deploying multiple subnets. Once-for-All~\cite{cai2020once} uses a progressive training supernet. BigNAS~\cite{yu2020bignas} uses a series of simple and practical training methods to improve the efficiency of training supernet. Once the supernet is trained, typical search algorithms, such as genetic search, can be applied to find a set of Pareto-optimal networks for various deployment scenarios.

In this work, we propose a unified depth pruning approach for both efficient CNN and vision transformer models with a progressive training strategy, a novel block pruning method, and the reparameterization technique. Differing from DepthShrinker and Layer-Folding finetuning the subnet with direct activation layer removal, our method progressively removes the activation layer in the pruned block during subnet training. Moreover, our method handles the normalization layer problem that DepthShrinker and Layer-Folding cannot prune models with LayerNorm or GroupNorm layers in the block. Further, they can not prune vision transformer models.
Although our work has a similar training process as VanillaNet~\cite{chen2023vanillanet}, whereas VanillaNet is proposed to design a completely new network structure, our method is a  general depth pruning framework for both CNN and vision transformer models. 

\section{Method}
\subsection{Unified Progressive Depth Pruner}
 
\begin{figure*}[t!]
    \centering
    \includegraphics[width=0.8\textwidth]{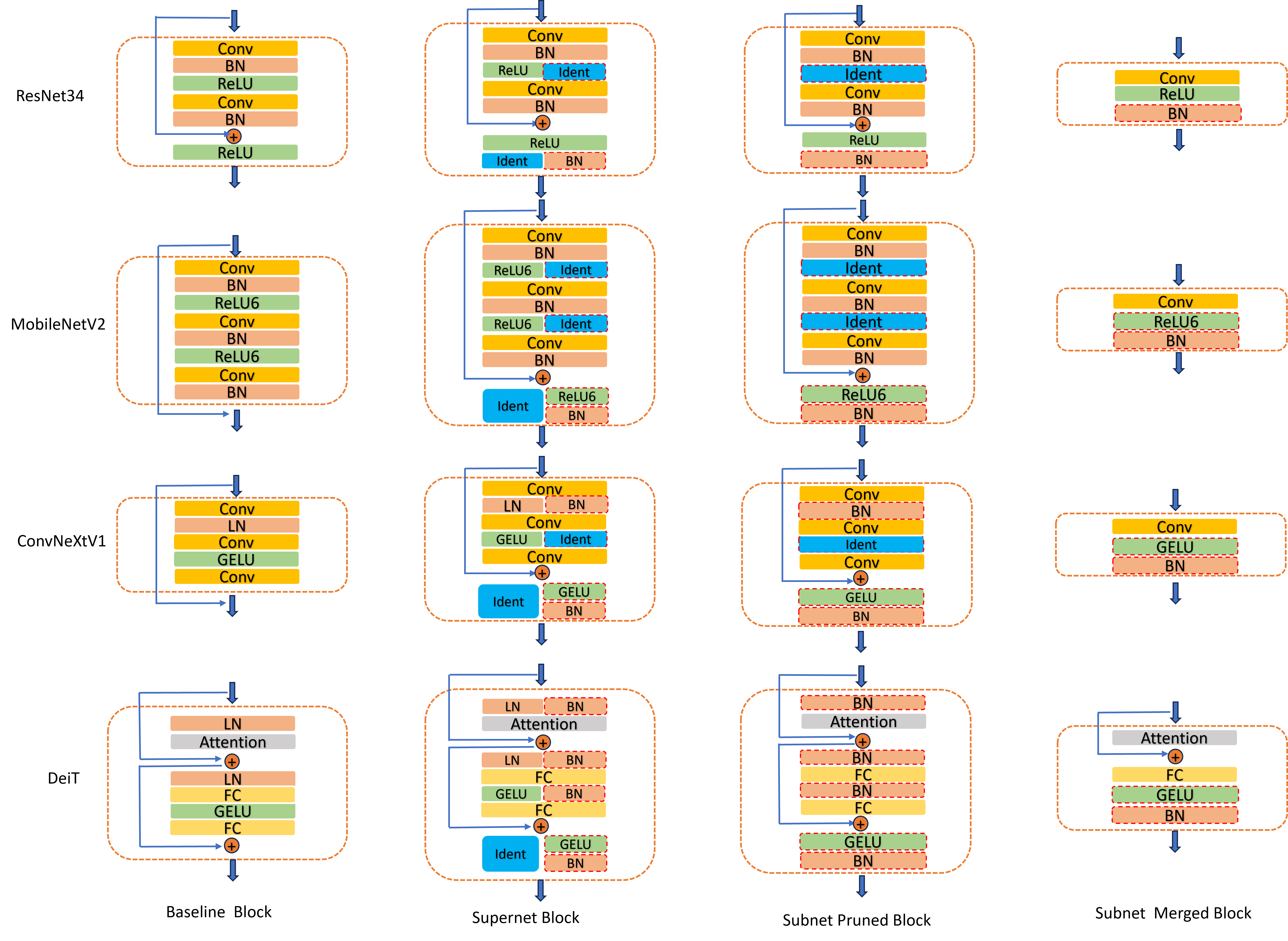}
   
    \caption{Framework overview of our proposed depth pruner. Each pruned baseline block will gradually evolve into a smaller merged block to speedup and save memory. Four baselines are experimented, including three CNN-based networks (ResNet34, MobileNetV2 and ConvNeXtV1) and one vision transformer network (DeiT-Tiny).}
    
    
    \label{fig:overall}
   
\end{figure*}

Our depth pruning approach aims to reduce model depth by proposed novel block pruning strategy with reparameterization technique rather than directly omitting the block. 
As shown in Figure~\ref{fig:overall}, our block pruning strategy converts a complex and slow block into a simple and fast block in block merging. For a block, we replace the activation layer with identity layer and replace the LayerNorm (LN) or GroupNorm (GN) layer with a BatchNorm (BN) layer and insert an activation layer with a BatchNorm layer at the end of block to create conditions for reparameterization. Then, the reparameterization technique can merge the BatchNorm layers,  
 adjacent Convolutional or Full-connection layers 
 and skip connections 
as shown in Figure~\ref{fig:overall}.

\textbf{Overview.} Our approach primarily consists of four main steps, which are supernet training, subnet searching, subnet training, and subnet merging. First, We construct a supernet based on the baseline model, where we make block modification as shown in Figure~\ref{fig:overall}. After supernet training, a search algorithm is used to search an optimal subnet. Then, we adopt a proposed progressive training strategy to optimize the optimal subnet with less accuracy loss. In the end, the subnet would be merged into a shallower model with the reparameterization technique.



\textbf{Supernet Training.} Efficient CNN and vision transformer models usually consist of several basic blocks, like some efficient models structure shown in Figure~\ref{fig:overall}. First, we  construct a supernet based on the baseline model and then train a robust supernet model based on the sandwich rule~\cite{yu2019universally} method to ensure that each subnet has a meaningful accuracy. We combine the baseline block and corresponding pruned block into a supernet block, which has both baseline  block and pruned block flows. For a supernet block, choosing the baseline block flow means no pruning and choosing pruned block flow means pruning the block. Then, subnet selection is a series of choices where the choice number is equal to a block number of the baseline model. The subnet would be faster with more pruned blocks selected.

Inspired by BigNAS~\cite{yu2020bignas}, we adopt the sandwich rule to sample the subnetwork before each step. In each step, we sample four sequential subnets, where the first one keeps all blocks unpruned, the following two subnets randomly select blocks to be pruned, and the last one keeps all blocks pruned. Then, we optimize supernet with accumulated grads of four subnets. The sandwich rule can effectively guarantee the upper and lower limits of the trained supernet. Also many methods~\cite{wang2021attentivenas,wang2021alphanet} demonstrate that the sandwich rule can  be used to train supernet efficiently, even if the number of epochs is small and the accuracy distribution of the subnet is the same as that of training more epochs. In this way, we can reduce the training cost of supernet. 

\textbf{Subnet Searching.} 
The primary objective of our depth pruner is to identify an optimal subnet based on a specified pruning criteria, such as the number of blocks to be pruned. As shown in Equation~\ref{eq:subnet_searching}, we formulate this problem as an optimal problem. For all samples $X$ and their labels $Y$, the goal of subnet searching is to find a subnet $S_p$ with the highest accuracy. $p \in R^{N_{block}}$ is a binary vector, representing the pruning setting of the subnet. If $i$-th block is pruned, $p_i$ is set to $1$. The number of pruned blocks of each subnet is equal to $k$. Genetic algorithm~\cite{cai2020once} is applied to solve this problem.

\begin{equation} 
\label{eq:subnet_searching}
\mathop{\arg\max} \limits_{p} {\text{Accuracy}(S_p(X), Y) \quad  \text{s.t.} \quad || p ||_0  = k}
\end{equation}

After the search process, we obtain a subnet that has a specified number of pruned blocks, and other blocks keep the same as the baseline model.

\textbf{Subnet Training.} We need to train the optimal subnet obtained from the previous step to restore its accuracy.  Rather than directly training the subnet, our approach employs a progressive training strategy to finetune the subnet smoothly transferring from baseline model weights.
The subnet training consists of two stages. During the first training stage, we adopt a progressive training strategy, gradually transferring from a baseline model structure to the pruned subnet structure by a controlling $\lambda$ factor. In the second stage, we continue finetuning the pruned subnet to the end for high accuracy.

As for the first stage, the gradual transition of $\lambda$ from 0 to 1 allows for a controlled process that transfers a baseline block to pruned block as shown in Equation~\ref{eq:subnet_training_lambda}, where $B_b$ is the block of baseline model and $B_p$ is the pruned block of the subnet and $x$ is the input of block. Thus all the pruned blocks of the subnet go through the same process with $\lambda$ and obtain the pruned subnet smoothly. 

\begin{equation} 
\label{eq:subnet_training_lambda}
  o =(1-\lambda)\cdot B_b(x)+ \lambda \cdot  B_p(x) 
\end{equation}
Our method  controls $\lambda$ transition from 0 to 1 during the first training phase, and then keeps $\lambda$ constant during the second phase of training as shown in Equation~\ref{eq:subnet_training_lambda_gen}, where $K$ is hyper-parameter and $C$ is current training epoch and $T$ is total training epoch. 
\begin{equation} 
\label{eq:subnet_training_lambda_gen}
\lambda = \begin{cases}
               1-max(0,cos(\frac{C\cdot K}{T}\cdot \frac{\pi}{2})), & \text{if } C \leq \frac{T}{K}, \\
               1, & \text{if } C > \frac{T}{K}. 
           \end{cases}
\end{equation}
It is worth noting that, to reduce the error in subsequent subnet merging, it is necessary to modify the padding and stride of related convolutional layers in pruned blocks before subnet training. For example, accumulate all padding values which are not zero forward to the first convolutional layer of the block and set the padding values of the remaining convolutional layer to zero. Also, accumulate all stride values which are not equal to one backward to the last convolutional layer of the block and set the stride value of other convolutional layers to one.

\textbf{Subnet Merging.} After subnet training, we obtain a sub-net with some activation layers replaced with Identity layers, some LayerNorm layers replaced with BatchNorm layers with some activation layers, and BatchNorm inserted at the end of the pruned block. In this stage, we adopt reparameterization techniques to make the subnet shallower.
\begin{itemize}
    \item $Conv+BN \longrightarrow Conv$
\end{itemize}
During the inference phase of a neural network, it is possible to fuse the operations of BatchNorm layers into Convolutional (Conv) layers to accelerate model inference. We assume that the parameters of the Conv layer are denoted as  $\omega$ and $b$, and the parameters of the BN layer are denoted as $\gamma,\sigma,\epsilon,\beta$. After merging of the Conv layer and BN layer, the parameters of the Conv layer would be modified as follows:
\begin{equation} 
\label{eq:merge_conv_bn}
\hat{\omega}=\frac{\gamma\cdot\omega}{\sqrt{\sigma^2+\epsilon}} \quad \hat{b}=\beta+\gamma\cdot\frac{b-\mu}{\sqrt{\sigma^2+\epsilon}}
\end{equation} 

\begin{itemize}
    \item $Conv/FC + Conv/FC \longrightarrow Conv/FC$
\end{itemize}

Two adjacent full-connection (FC) layers can be simply merged into a FC layer by the fusion of their weights. Suppose that there are two adjacent FC layers, their weights are $W_1 \in R^{C_1 \times C_0}$ and $W_2 \in R^{C_2 \times C_1}$, and their biases are $b_1 \in R^{C_1}$ and $b_2 \in R^{C_2}$. For a given input feature $x \in R^{N \times C_0}$, the output of these two FC layers is expressed as $W_2(W_1 x^T+b_1)+b_2$, which can be obtained by an equivalent FC layer whose weight is equal to $W_2 W_1$ and bias is equal to $(W_2 b_1 + b_2)$.

We will primarily introduce the fusion methods between convolutional layers. For sequential $1\times 1$  convolutional layer fusing with  $k\times k$  convolutional layer, we adopt the proposed fusion method from DBB ~\cite{ding2021diverse} to merge the two layers into an equivalent $k\times k$ convolutional layer. For sequential $k\times k$  convolutional layer fusing with  $k\times k$, it would obtain an equivalent $(2k-1)\times (2k-1)$ convolutional layer with more parameters and flops compared with $k\times k$ convolutional layer. We propose a simple way to address this problem, which 
replaces the first $k\times k$ Conv layer with a $1\times 1$ Conv layer, where the parameter of $1\times 1$ Conv is taken from the central point of the  $k\times k$ Conv with a retraining model. Then take the DBB fusion method to transform the two convolutional layers into an equivalent $k\times k$ convolutional layer.

\begin{itemize}
    \item $Conv + skip\text{ }connection(Identity/Conv) \longrightarrow Conv$
\end{itemize}
RepVGG~\cite{ding2021repvgg} proposes a method to transform a multi-branch model into an equivalent single-path model. According to the additivity property of convolutions, for two convolutional layers with the same kernel size, they satisfy the following Equation~\ref{eq:merge_convs}:
\begin{equation} 
\label{eq:merge_convs}
Conv(W_1, x) + Conv(W_2, x) = Conv(W_1 + W_2, x)
\end{equation}
where $Conv(W, x)$ represents the convolutional operation, $W_1$ and $W_2$ are the convolutional kernel parameters, and $x$ is the input data. RepVGG has demonstrated that the identity and $1\times 1$ convolutional can be equivalently transformed into a $k\times k$ convolution. Then with the property of convolutional additivity, the multi-branch convolutional layers can be merged into an equivalent convolutional layer and skip connection identity can be merged into convolutional layer too.





\subsection{Depth Pruner on CNN}

Applying our method on CNN models can refer to Figure~\ref{fig:overall} showing the pipeline. We should find the basic block first, and design a corresponding pruned block by reference of the pruned block in Figure~\ref{fig:overall}. For activation layers in the block, we replace it with an Identity layer. For the normalization layer, which is not BatchNorm layer in block, we replace it with a BatchNorm layer, otherwise nothing needs to be done. Finally, we would insert an activation layer with a BatchNorm layer at the end of block. If an activation layer already exists in the position like ResNet34 block, only a BatchNorm layer needs be inserted after the activation layer at the end of block.  After the pruned block is completed, review the supernet training, subent searching, subnet training, and subnet merging processes. We would obtain the pruned CNN models. For plain CNN models, we can define the block which can includes two or more sequential convolutional layers.



\subsection{Depth Pruner on Vision Transformer}
We also apply our proposed depth pruner on vision transformer models. The vision transformer block usually has a multi-headed self-attention (MHSA) module and a MLP module that includes two full-connection layers. Particularly, we utilize DeiT~\cite{deit2021} as the case showing the pruning flows. As demonstrated in Figure~\ref{fig:overall}, to build the Supernet, we add BN bypasses next to the LN and activation (GELU) layers of the original model and insert a GELU\&BN block after the residual addition operation. After subnet searching and subnet training, we obtain the subnet, whose original LN and GELU operations of the pruned blocks are all replaced by BNs. A GELU\&BN block is attached after the residual addition. Then, we merge the subnet to obtain a fast pruned model as shown in Figure~\ref{fig:overall}. 

\section{Experiments}
In this section, we showcase the efficacy of our depth pruner. Initially, we elucidate the experimental configurations and outline the procedure for applying the depth pruner to both CNN models and Vision Transformers. Subsequently, we compare our results with the state-of-the-art pruning methods to highlight the superiority of our approach. Finally, we perform ablation studies to elaborate on the effect of subnet searching and progressive training strategy in our method.
\subsection{Datasets}
All the experiments are conducted on the ImageNet-1K~\cite{russakovsky2015imagenet}. ImageNet-1K dataset is a widely used image classification dataset that spans 1000 object classes and contains 1,281,167 training images, 50,000 validation images, and 100,000 test images. We apply conventional data augmentation techniques to preprocess input images during training and scale input images to $224 \times 224$ for all experiments with reporting performance on validation dataset.
\subsection{Experiments Setting on Different Models}
We apply depth pruner on a series of CNN models, including ResNet34~\cite{he2016deep}, MobileNetV2, ConvNeXtV1~\cite{liu2022convnet}, and Vision Transformer~\cite{deit2021} to validate the efficiency of our method. We utilize four GPUs to train our model, with a total batch size of 256. In the training process, we take 10 epochs to train the supernet, except for MobileNetV2 and search optimal subnets. Then we train these subnets with the proposed progressive training strategy and complete subnet merging  to obtain more efficient shallow models.

\textbf{ResNet34.} For ResNet34 pruning experiments, we  prune 6 and 10 blocks respectively and go through the whole pruning process to obtain two pruned and shallow subnets. For subnet training, the hyper-parameters $K$ in Equation~\ref{eq:subnet_training_lambda_gen} is 3, and the total training epochs is 150. At epoch 100, we change kernel size from $3\times3$ to $1\times1$ of the first convolutional layer in the pruned block. We compare our method with MetaPruning~\cite{liu2019metapruning} and channel-wise NAS method Universally Slimmable Networks (US) ~\cite{yu2019universally} to verify the pruning performance.

\textbf{MobileNetV2.} For MobileNetV2 pruning experiments, we adopt three pruning configurations from DepthShrinker. We skip the supernet training and subnet searching phases to obtain three same subnets. We directly train these three subnets to the end and compare the final performance with the corresponding subnet. We also use the NAS method to search three subnets with similar speedup ratios to compare with our method. For subnet training, the hyper-parameters $K$  is 3, and the total training epochs is 450.

\begin{table}[t]
\centering
\begin{tabular}{c|c|c|c}
\toprule  
Models& \makecell[c]{FLOPs\\(G)} & \makecell[c]{Acc1\\(\%)}& Speedup\\
\midrule  
ResNet34-Baseline &  3.67 & 73.6 & 1.00 \\
\midrule  
US-ResNet34-0.6$\times$ & 2.32& 72.0 & 1.24 \\
MetaPruning-0.6$\times$ & 2.32 & 72.8 & 1.24 \\
\textbf{Ours-P6} & 2.97 &  \textbf{73.2}& \textbf{1.25} \\
\textbf{Ours-P6*} & 2.97 & \textbf{73.5}& \textbf{1.24} \\
\midrule
US-ResNet34-0.5$\times$ & 1.90 & 70.8 & 1.43 \\
MetaPruning-0.5$\times$ & 1.87& 71.6 & 1.43 \\
\textbf{Ours-P10} & 2.51 & \textbf{71.8}& \textbf{1.43} \\
\textbf{Ours-P10*} & 2.51 &  \textbf{72.4}& \textbf{1.43} \\
\bottomrule 
\end{tabular}
\caption{Classification performance comparisons on ImageNet. P6 and P10 indicate pruning 6 and 10 blocks, respectively, and ’*’ means a higher accuracy subnet with longer search time.}

\label{tab:r34}
\end{table}

\textbf{ConvNeXtV1.} For ConvNeXtV1 pruning experiments, we set three pruning configurations which let the subnet obtain the performance around 81\% top-1 accuracy on ImageNet-1k. Then, we compare these subnets with many SOTA models which include CNN and vision transformer structures to verify our method pruning performance.  For subnet training, the hyper-parameters $K$ is 4.5 and the total training epochs is 450. To achieve better speedup, we change the kernel of the depthwise conv of the pruned block from 7 to 3.

\textbf{DeiT.} For DeiT pruning experiments, we conduct a pruning experiment with 6 layers experiment to compare with the SOTA vision transformer pruning method. For subnet training, the hyper-parameters $K$  is 6, and the total training epochs is 450. 

\subsection{Comparisons with SOTA Models}
We compare the depth pruner with the state-of-the-art pruning methods under the comparable inference speed on a single AMD MI100 GPU. Following~\cite{graham2021levit}, we measure the average inference speedup of compressed networks with batchsize=128. In this paper, we compare the accuracy of different models with comparable speedups.

\textbf{ResNet34.} Table~\ref{tab:r34}  compares our method with MetaPruning and NAS US methods on ResNet34. We prune 6 and 10 blocks respectively by applying our depth pruner to obtain two subnets with 1.25$\times$ and 1.43$\times$ speedup ratios, respectively. Under comparable speedup, our method surpasses the MetaPruing by 0.8\% and NAS US method by 1.6\% on 1.43$\times$ speedup ratio with a longer search time.

\begin{table}[t]
\centering
\begin{tabular}{c|c|c|c}
\toprule  
Models& \makecell[c]{FLOPs \\(M)}& \makecell[c]{Acc1\\(\%)}& Speedup \\
\midrule  
MBV2-1.4-Baseline & 630 & 76.5 & 1.00\\
\midrule  
MetaPruning-0.5$\times$ & 332 & 73.2 & 1.49\\
US-MBV2-1.4-0.6$\times$ & 384 & 73.4& 1.56\\
MBV2-1.4-DS-A & 519& 74.4& 1.75\\
\textbf{Ours-P6} &  519 & \textbf{74.8} & \textbf{1.75} \\
\midrule
US-MBV2-1.4-0.4$\times$ & 286& 72.2& 2.04\\
MBV2-1.4-DS-C & 492& 73.1& 2.16\\
\textbf{Ours-P9 }& 492 &\textbf{73.8}& \textbf{2.16} \\
\midrule
US-MBV2-1.4-0.3$\times$ & 213 & 68.1 & 2.40\\
MBV2-1.4-DS-E & 474& 72.2& 2.50\\
\textbf{Ours-P11} & 474 &\textbf{72.5} & \textbf{2.50} \\
\bottomrule 
\end{tabular}
\caption{Classification performance comparisons with MetaPruing, NAS US and DepthShrinker on ImageNet with same network structures as the DepthShinker.}
\label{tab:mbv2}

\end{table}

\begin{table*}[t]
\centering
\begin{tabular}{c|c|c|c|c|c}
\toprule  
Models & Type & FLOPs(G) & Params(M)  & Acc1(\%) & Speedup\\
\midrule  


ConvNeXtV1-T-Baseline~\cite{liu2022convnet} & Conv & 4.5 & 28.6 & 82.1 & 2.4 \\
\midrule
RepVGG-B1~\cite{ding2021repvgg} & Conv & 11.8 & 51.8 & 78.4 & 2.1 \\
EfficientFormerV2-S1~\cite{li2022rethinking} & Hybrid & 0.7 & 6.1 & 79.0 & 4.0 \\
MobileOne-S4~\cite{vasu2023mobileone} & Conv & 3.0 & 14.8 & 79.4 & 3.9 \\
ConvNeXtV2-P~\cite{woo2023convnext} & Conv & 9.1 & 9.1 & 79.7 & 4.0 \\
PVT-Small~\cite{wang2021pyramid} & Attention & 3.8 & 24.5 & 79.8 & 2.7 \\
VanillaNet-9~\cite{chen2023vanillanet} & Conv & 8.6 & 41.4 & 79.9 & 1.1 \\
RegNetY-12G~\cite{radosavovic2020designing} & Conv & 12.1 & 51.8 & 80.3 & 1.8 \\
\rowcolor{lightgray} \textbf{ConvNeXtV1-T-UPDP-P9} & Conv & 2.5 & 23.6 & \textbf{80.6} & \textbf{4.9} \\
\midrule
RepVGG-B3~\cite{ding2021repvgg} & Conv & 26.2 & 110.9 & 80.6 & 1.1 \\
ConvNeXtV1-N~\cite{woo2023convnext} & Conv & 2.5 & 15.6 & 80.8 & 3.6 \\
EdgeViT-S~\cite{pan2022edgevits} & Hybrid & 1.9 & 11.1 & 81.0 & 3.6 \\
Swin-T~\cite{liu2021swin} & Attention & 4.5 & 28.3 & 81.1 & 2.8 \\
PVT-Medium~\cite{wang2021pyramid} & Attention & 6.7 & 44.0 & 81.2 & 1.8 \\
\rowcolor{lightgray} \textbf{ConvNeXtV1-T-UPDP-P6} & Conv & 3.1 & 27.5 & \textbf{81.3} & \textbf{4.2} \\
\midrule
VanillaNet-11~\cite{chen2023vanillanet} & Conv & 10.3 & 50.0 & 81.1 & 1.1 \\
EfficientFormerV2-S2~\cite{li2022rethinking} & Hybrid & 1.3 & 12.6 & 81.6 & 1.0 \\
PVT-Large~\cite{wang2021pyramid} & Attention & 9.8 & 61.0 & 81.7 & 1.3 \\
DeiT-B~\cite{deit2021} & Attention & 17.5 & 86.0 & 81.8 & 1.5 \\
\rowcolor{lightgray} \textbf{ConvNeXtV1-T-UPDP-P3} & Conv & 3.8 & 28.3 & \textbf{81.9} & \textbf{3.3} \\

\bottomrule 
\end{tabular}
\caption{Performance of ConvNeXtV1 depth pruning results on ImgeNet. Speedups are tested on an AMD MI100 GPU with a batch size of 128. Adopt the slowest network in the table (EfficientFormerV2) as the baseline(1.0 speedup) for comparison.}
\label{tab:convnextv1}

\end{table*}

\textbf{MobileNetV2.} Table~\ref{tab:mbv2} shows the experimental results on MobileNetV2-1.4. We adopt the same subnets as DepthShrinker, but the subnet training process is different from our progressive training strategy. We achieve 0.7\% higher accuracy than MBV2-1.4-DS-C at a 2.16$\times$ speedup ratio, and some improvement compared to DepthShrinker at other speedup ratios. We also compare MetaPruning, and similar to ResNet34, we reproduce MetaPruning-0.35$\times$  with inference speeds comparable to MBV2-1.4-DS-C, while our depth pruning achieves a 2.1\% higher accuracy with a higher speedup ratio.

\begin{table}[t]
\centering
\begin{tabular}{c|c|c|c|c}
\toprule  
Models& \makecell[c]{FLOPs\\ (G)} &  \makecell[c]{Params\\(M)}& \makecell[c]{Acc1\\(\%)} & Speedup \\
\midrule  
DeiT-Tiny & 1.3 & 5.4 & 72.2 & 1.00  \\
\midrule

SCOP* & 0.8 & - & 68.9 & - \\
HVT* &  0.7 & - & 69.7 & - \\

S$^2$ViTE & 1.0 & 4.2 & 70.1 & 1.12 \\ 
WD-Pruning & 0.7 & 3.5 & 70.3 & 1.20 \\
XPruner & 0.6 & -& 71.1 & -\\
\textbf{Ours-P6} & 0.9 & 3.8 & \textbf{70.3} & \textbf{1.26} \\
\bottomrule 
\end{tabular}
\caption{DeiT depth pruning results on ImageNet. The results of S$^2$ViTE \cite{tang2022svite} and WD-Pruning~\cite{wdpruner2022} refer to their paper. SCOP~\cite{2020_scop}, HVT~\cite{pan2021scalable}, and XPruner~\cite{Yu2023XPrunerEP} do not publish their results about the number of parameters and speedup ratio. "*" denotes that the results come from~\cite{wdpruner2022}.}
\label{tab:deit}

\end{table}

\textbf{ConvNeXtV1.} Table~\ref{tab:convnextv1} compares our accuracy with some common efficient models since there is no compression method for ConvNeXtV1. We test the speedup ratios of all networks on the AMD platform using the slowest network EfficientFormerV2-S2 in the table as a benchmark. We divide the model into levels by accuracy, and our depth pruning method achieves higher accuracy with comparable speed in different levels.

\textbf{DeiT.} As shown in Table~\ref{tab:deit}, our method outperforms other state-of-the-art methods in both accuracy and speedup ratio. Our proposed depth pruner achieves a 1.26$\times$ speedup ratio with only a 1.9\% top-1 accuracy drop. By replacing mergeable modules and applying the reparameterization technique, our proposed method can shrink the network and bring real inference acceleration. 

\subsection{Ablation Study}
In this section, we analyze the effectiveness of subnet searching and progressive training strategy.

\begin{table}[H]
\centering
\begin{tabular}{c|c|c}
\toprule  
Models& Before FT(\%)& After FT(\%)\\
\midrule  
\textbf{ResNet34-P10-A} & \textbf{57.8}& \textbf{71.8} \\
ResNet34-P10-B & 55.9 & 71.2 \\
\bottomrule 
\end{tabular}
\caption{Evaluating the accuracy consistence of subnets.}
\label{tab:search}

\end{table}

\textbf{Effectiveness of Subnet Searching.}
We verify the effectiveness of our ResNet34 subnet searching by comparing performance of two pruned-10-layers subnets with different accuracy before subnet finetune (FT) based on ResNet34. Table~\ref{tab:search} shows ResNet34-P10-A with a higher accuracy before subnet finetune can achieve higher finetune accuracy, which proves the effectiveness of supernet training and subnet searching for optimal subnet with a final high performance.

\begin{table}[H]
\centering
\begin{tabular}{c|c|c}
\toprule  
Models& Direct (\%) & Progressive (\%)\\
\midrule  
ResNet34-P10 & 71.3& \textbf{71.8} \\
MBV2-1.4-P9 & 73.1 & \textbf{73.8} \\
ConvNeXtV1-T-P3  & 81.6 & \textbf{81.9} \\
DeiT-Tiny-P6 & 69.5 & \textbf{70.3} \\
\bottomrule 
\end{tabular}
\caption{Evaluating the effectiveness of progressive training.}
\label{tab:progressive}

\end{table}

\textbf{Effectiveness of Progressive Training Strategy.} Compared with hard removal of non-linear activation functions, our progressive training has a significant improvement in the accuracy of each subnetwork. As shown in Table~\ref{tab:progressive}, for various sub-networks, we observe that progressive training improves accuracy by 0.3\%-0.8\% than direct training method.



\section{Conclusion}
In this paper, we present a unified depth pruner for both efficient CNN and vision transformer models to prune models in the depth dimension. Our depth pruner includes four steps, which are supernet training, subnet searching, subnet training, and subnet merging. We propose a novel block pruning method and a progressive training strategy to utilize baseline model weights better. During subnet merging, we use reparameterization technique to make subnet become shallower and faster. We conduct our method to several CNN models and transformer models. The SOTA pruning performance demonstrates the superiority of our method. In the future, we would explore our method on more transformer models and tasks.

\bibliography{depth_pruner}

\end{document}